\documentclass{article}

\usepackage{microtype}
\usepackage{graphicx}
\usepackage{subfigure}
\usepackage{booktabs} 

\usepackage{hyperref}

\usepackage[accepted]{icml2024}

\usepackage{amsmath}
\usepackage{amssymb}
\usepackage{mathtools}
\usepackage{amsthm}

\usepackage[capitalize,noabbrev]{cleveref}

\usepackage{times}
\usepackage{epsfig}
\usepackage{graphicx}

\theoremstyle{plain}

\theoremstyle{definition}

\theoremstyle{remark}

\usepackage[textsize=tiny]{todonotes}
\newcommand{\ie}{i.e.,}
\newcommand{\vs}{v.s.,}
\newcommand{\eg}{e.g.,}

\icmltitlerunning{Physics-Enhanced Multi-fidelity Learning for Optical Surface Imprint}

\begin{document}

\twocolumn[
\icmltitle{Physics-Enhanced Multi-fidelity Learning for Optical Surface Imprint}





\begin{icmlauthorlist}
\icmlauthor{Yongchao Chen}{yyy,comp}
\end{icmlauthorlist}

\icmlaffiliation{yyy}{Massachusetts Institute of Technology, Boston MA, USA}
\icmlaffiliation{comp}{Harvard University, Boston MA, USA}

\icmlcorrespondingauthor{Yongchao Chen}{yongchaochen@fas.harvard.edu}

\icmlkeywords{Machine Learning, ICML}

\vskip 0.3in
]



\printAffiliationsAndNotice{} 

\begin{abstract}
   Human fingerprints serve as one unique and powerful characteristic for each person, from which policemen can recognize the identity. Similar to humans, many natural bodies and intrinsic mechanical qualities can also be uniquely identified from surface characteristics. To measure the elasto-plastic properties of one material, one formally sharp indenter is pushed into the measured body under constant force and retracted, leaving a unique residual imprint of the minute size from several micrometers to nanometers. However, one great challenge is how to map the optical image of this residual imprint into the real wanted mechanical properties, \ie, the tensile force curve. In this paper, we propose a novel method to use multi-fidelity neural networks (MFNN) to solve this inverse problem. We first build up the NN model via pure simulation data, and then bridge the sim-to-real gap via transfer learning. Considering the difficulty of collecting real experimental data, we use NN to dig out the unknown physics and also implant the known physics into the transfer learning framework, thus highly improving the model stability and decreasing the data requirement. The final constructed model only needs three-shot calibration of real materials. We tested the final model across 20 real materials and achieved satisfying accuracy. This work serves as one great example of applying machine learning into scientific research, especially under the constraints of data limitation and fidelity variance.
\end{abstract}

\section{Introduction}
\label{sec:intro}

Over the past century, humans have been searching for the optimal natural or artificial materials with most suitable mechanical properties. To accelerate this searching process, persistent research has focused on developing platforms for high-throughout (HT) synthesis and characterization of materials~\cite{ament2021autonomous, erps2021accelerated}. Benefiting from its intrinsic experimental simplicity and broad applicability, indentation has been considered as a paradigm for HT probe of mechanical properties of materials~\cite{chen2021tuning,lu2020extraction}. Indentation requires minimal specimen preparation and mounting, rewards hundreds of data from a single specimen, and can probe materials across from nano to macro scales via varied loads~\cite{doerner1986method,chen2022anomalous}. Using one sharp indenter to push into the material surface under the constant load, the material surface will be crashed to form a crown-like pattern which contains plentiful information to reveal elasto-plastic properties of materials. However, since the inverse mapping from the crashed pattern into formal material parameters is quite complex, many former researchers regard the pattern information as a rough criterion~\cite{jeong2021evaluation}. On the other hand, compared to the traditional characterization methods like tensile testing, direct measurement of optical profile is more convenient. Optical measurement has been widely applied in many areas, such as biological systems~\cite{cooke2021physics,smith2018tracking}, particle tracking~\cite{zalevsky2009simultaneous} and vibrations~\cite{cuomo2022scientific,zhong2021towards}.

The rapid development in artificial intelligence has aroused one great opportunity of AI for science~\cite{taddeo2018ai,degrave2022magnetic,fawzi2022discovering}. One big limitation is that the data in many scientific areas are not plentiful, especially for high-fidelity experimental data. To overcome this problem, many new machine learning frameworks incorporating physical constraints have been proposed, \ie, physics-informed neural networks (PINN)~\cite{cuomo2022scientific,raissi2019physics}. To decrease the requirements of real experimental data, some pre-training and transfer learning frameworks have been proposed. One rough model will be trained with many simulation data and then fine-tuned by some experimental data. In this way, the requirements of high-fidelity data will be decreased. Another big issue in scientific AI is that many problems in scientific areas are inverse problems, \ie, the input-output relation may be ill-conditioned~\cite{tarantola2005inverse,meng2017insight,Song_2023_WACV,hu2023deep}. Sometimes, different input parameters correspond to the similar or even the same output results, thus making the NN prediction to be inaccurate when trying to infer the results back to the inputs. This is referred as the non-unique problem.

Herein, we attempt to optimize the inverse mapping from the residual pattern formed by indentation into the formal stress-strain curve of materials through MFNN. To make sure our problem is not ill-conditioned so that the model can predict well, we first apply NN to explore the forward problem and then combine the optimization method to search the possibility of non-uniqueness. This method assists to determine what suitable features to choose. Our scientific discovery is that instead of choosing the load-displacement curve, the residual pattern is already informative enough for predicting stress-strain relation. This conclusion turns out to be consistent with former mechanical theories of strain fields.

As for the NN architecture, the whole process is achieved by first building up an initial NN model based on a large amount of 2D simulation data through finite element methods (FEM), and then transferring it into the 3D model using some 3D FEM simulation data. The transferred model gets further fine-tuned by incorporating some real experimental data and corresponding physical constraints into the model. To make the transfer learning process more efficient and stabilized, two physical parameters (friction coefficient $\mu$ and Poisson's ratio $\nu$) are tuned in the simulation set, resulting into an ensemble of three parallel NNs. The final model is constructed from these NNs. It turns out that this step of tuning physical parameters is quite critical to the prediction accuracy. The underlying principle is owing to the closer sim-to-real gap after tuning the friction coefficient $\mu$ and Poisson's ratio $\nu$. The final result owns satisfying accuracy when predicting the real stress-strain relation of the testing materials. We further notice that the initial pre-assumed material model to describe material behaviors will greatly influence the final prediction result. Hence, a method without pre-assuming material model is proposed and achieves relatively great accuracy. In summary, the main contributions of the paper include:
\begin{itemize}
  \item To the best of our knowledge, the first attempt to apply NN framework to inverting the optical profile of the indentation imprint to the real material elasto-plastic properties. The models with and without pre-defined mechanical laws are respectively trained and discussed.
  \item Designing the MFNN framework to combine multi-fidelity data from 2D and 3D FEM simulation, and real experiments. Incorporating the physical intuitions into the MFNN model to decrease the data requirements and stabilize the model. The constructed model only needs real experimental data of 3 types of materials. We also contributed experimental results of other 20 types of materials. It turns out that our model achieved an average 3.4\% relative error across 20 testing real materials.
  \item Applying forward NN and BFGS~\cite{yuan1991modified} optimization to explore the non-unique issue in the inverse problem and dig out the required features for training. Approaches based on NN to approximate forward models are a powerful tool that have recently seen more use in computational imaging. Our work supplement that using such an approach to speed up and differentiate FEM is an effective application.
  \item Release of the dataset and code for indentation and corresponding residual imprint features. Our MFNN framework can serve as a benchmark in this problem.
\end{itemize}

\begin{figure}[t]
  \centering
   \includegraphics[width=1\linewidth]{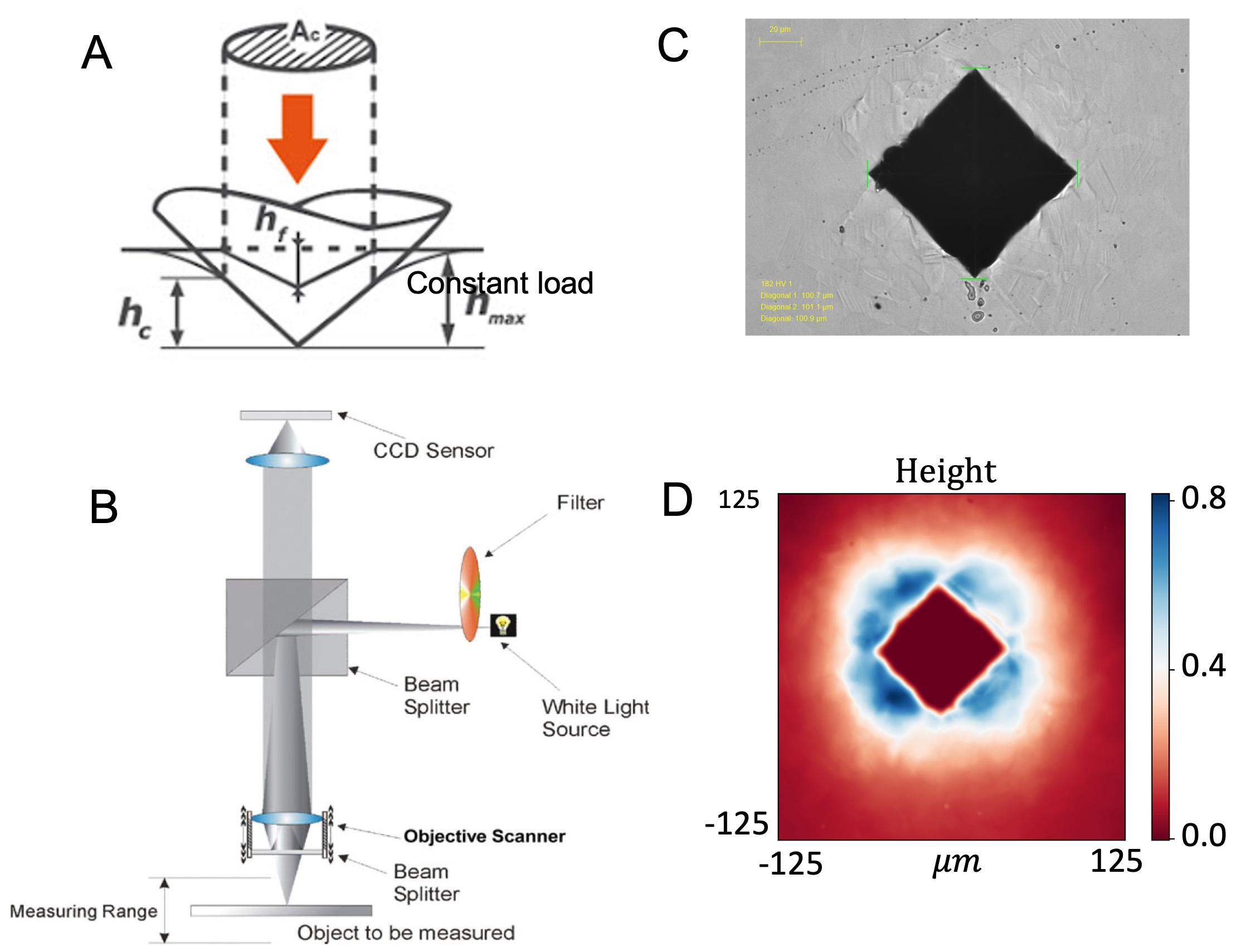}

   \caption{
Experimental methods. (A and B) Schematic illustrations of indentation and optical profilometer, respectively. (C) Typical pile-up image taken from scanning electron microscope. (D) Typical height distribution image measured by optical profilometer.}
   \label{fig:onefigure}
\end{figure}

\section{Backgrounds and methods}
\label{sec:back}

\subsection{Indentation}
\begin{figure*}[t]
  \centering
   \includegraphics[width=0.65\linewidth]{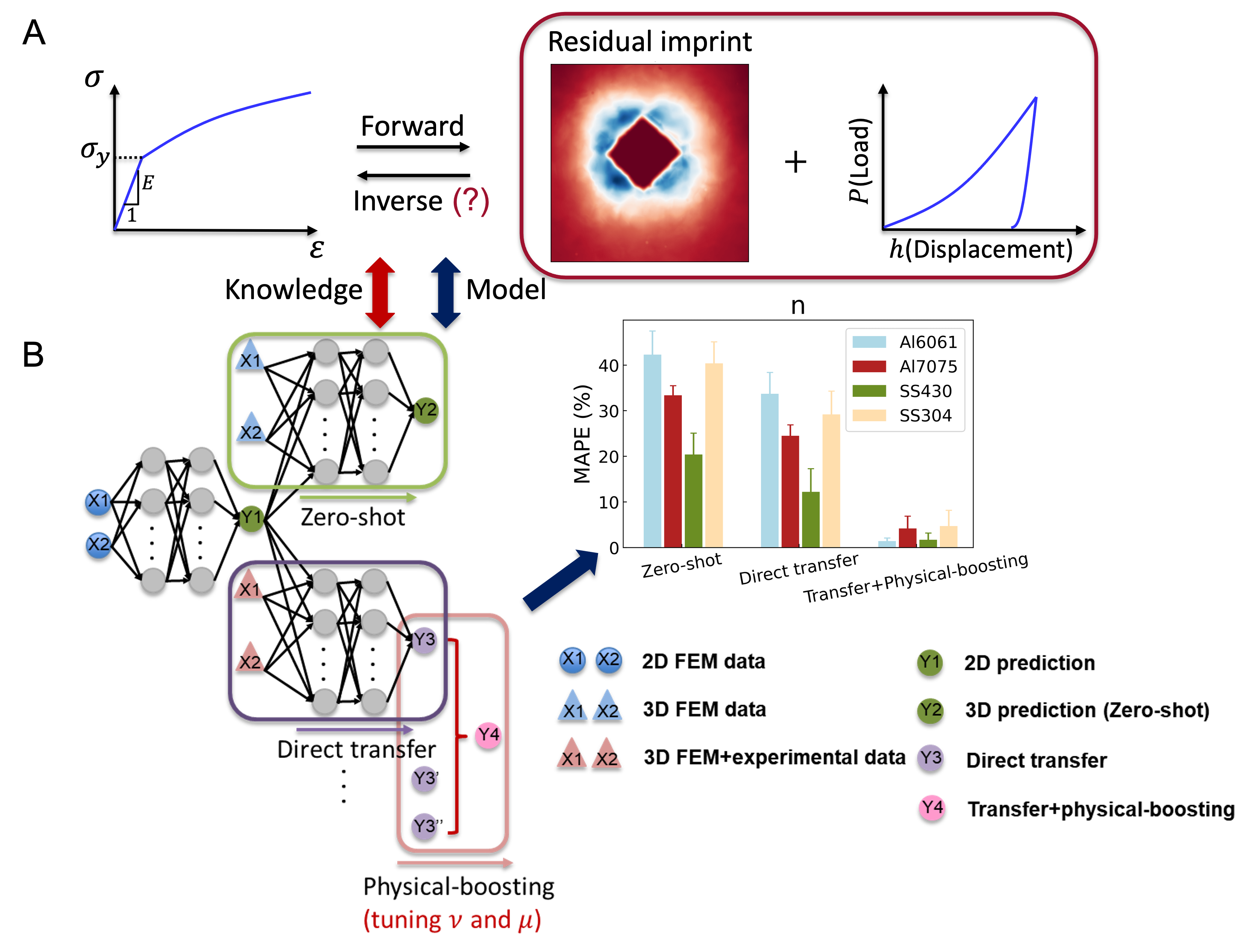}

   \caption{Transfer learning to solve the indentation inverse problem via residual imprint (pile-up). (A) Schematic illustration of indentation forward and inverse problems. Materials conforming to typical hardening behaviors (left) will form pile-up on sample surfaces after indentation and response typical load-displacement curves (right). (B) Flowcharts of the transfer learning DNN employed in this study.}
   \label{fig:twofigure}
\end{figure*}

As shown in \cref{fig:onefigure}A, one formal-shaped indenter (here is four-fold) is pushed into the material surface to create one minute crater (also called pile-up) (\cref{fig:onefigure}C). There are two types of indentation techniques. The first technique is to apply the constant load ($P$) and only use the division of load ($P$) and projection area ($A$) of the crater to describe material's property. That is called hardness ($H$).
\begin{equation}
  H = \dfrac{P}{A}
  \label{eq:important}
\end{equation}
While this method is convenient, the final measured result $H$ is not directly related to material parameters, \ie, the stress-strain relation. To acquire more information, the second method records the load evolution($P$) \vs indenting depth($h$) (\cref{fig:twofigure}A right) and tries to map from the load-depth curve into material stress-strain relation ((\cref{fig:twofigure}A left). However, many former research reveal that the load-depth curve may not be informative enough~\cite{chen2007uniqueness,campbell2018experimental}. Here we show that the first technique (hardness measuring) plus optical profile can already make the inverse problem well-conditioned, even without the need for load-depth curve.
\subsection{Optical profilometer}
Our goal is to utilize the optical profilometer~\cite{fainman1982optical} to measure the surface height map of the residual imprint. Specifically, as shown in \cref{fig:onefigure}B, a beam of light from a single source is split by the interferometer into two separate beams. Each of these beams travel separate paths, one onto a reference surface and the other onto the surface to be measured. The beams are then recombined resulting in an interference pattern. An imaging device, usually a CCD array, is used to collect this information. By moving the interferometer vertically away from the measurement surface, the point at which this interference occurs can be found for each pixel of the CCD. By tracking the position of the interferometer during this process a 3D map (\cref{fig:onefigure}D) of the surface can be formed.

\subsection{Forward and inverse problem}
\Cref{fig:twofigure} schematically illustrates the problem set of our study. \cref{fig:twofigure}A (left side) is a schematic diagram showing a typical stress-strain response of a power-law strain-hardening material which can be used for many engineering metallic materials. The elastic behavior follows Hook’s law, whereas the plastic response is approximated by different constitutive models~\cite{hertele2011generic}. One assumption is the three-parameter Hollomon model (fitting parameters: $E$, $\sigma_y$, $n$), in which true stress $\sigma$ and true strain $\varepsilon$ are related as:
\begin{equation}
\sigma=
    \begin{cases}
        E\varepsilon,\quad\sigma<\sigma_y\\
        E\varepsilon_y^{1-n}\varepsilon^n,\quad\sigma\geq\sigma_y
    \end{cases}
\end{equation}
, while another assuming model is the four-parameter Ludwik model (fitting parameters: $E$, $\sigma_y$, $n$, $K$), displayed as:
\begin{equation}
\sigma=
    \begin{cases}
        E\varepsilon,\quad\sigma<\sigma_y\\
        K\varepsilon_p^n,\quad\sigma\geq\sigma_y
    \end{cases}
\end{equation}
, where E is the elastic modulus, $\sigma_y$ is the yield stress, $K$ is the work hardening coefficient, n is the work hardening exponent, and $\varepsilon_p$ is the equivalent plastic strain determined as $\varepsilon_p=\varepsilon-\dfrac{\sigma}{E}$. We find that the pre-assumed model sometimes fails to accurately fit the stress-strain curves measured in the real experiments, as shown in Appendix~\ref{sec:appendix1}. Later we will also turn into the study without pre-assuming the constitutive model and represent stress-strain curves via point-to-point linear connection.

Knowing the stress-strain relation of one material, theoretically we can uniquely determine the residual imprint and load-depth curve. This is referred as the forward problem. However, how to inversely determine the stress-strain curve from the residual imprint remains challenging.
\subsection{FEM simulation}
To simulate the elasto-plastic behaviors of the system, the ABAQUS (Dassault Systèmes Simulia Corp.) software package is employed to conduct 2D and 3D FEM analysis~\cite{reddy2019introduction}. For single 16-core CPU, each 2D job costs around 10 minutes, while each 3D job costs 6 hours.
\subsection{Network design and dataset construction}
Specific NN structure is displayed in \cref{fig:twofigure}B,comprising several independent NNs connected by extra parameters. Each NN owns 6 hidden layers capable of learning the variations of pile-up features and hardness with respect to different stress-strain properties. Each hidden layer owns 32 neurons. ReLU is used as the activation function. The Adam optimizer~\cite{kingma2014adam} is applied in training by setting the learning rate as 0.0001. Appendix~\ref{sec:appendix2} shows the dataset statistics, in which the parameters of FEM materials are wide enough to nearly cover all the metallic materials. We mainly use the dataset of Ludwik model to train the NNs in the following sections.
\section{Related work}
\label{sec:rela}
With the great progress of computer vision~\cite{he2016deep,shorten2019survey,hu2023fedssc}, combining optical microscope with NN to infer underlying properties is becoming more and more prevalent~\cite{sheinin2022dual,cooke2021physics}. As for inferring elasto-plastic properties from residual imprint, the traditional methods mainly focused on doing FEM iteration to match the simulation results with experiments~\cite{campbell2018experimental,meng2017insight,jeong2021evaluation}. These methods encounter two issues that 1) The iterations of FEM to acquire the best set of parameters to fit experimental results will consume much time; 2) The sim-to-real gap is not fixed, thus the predicted parameters may encounter great errors in some cases. In recent years, some researchers have tried to apply NN to solve this inverse problem~\cite{lu2020extraction,haj2008nonlinear,jeong2021evaluation}. However, they all directly utilized the features of load-depth curves without the features of residual imprints, which may encounter unique problem. Meanwhile, the relation between the input features and the predicted elastic and plastic properties are not well revealed or explained by machine learning. In other words, more efforts are needed to utilize machine learning to help explore the underlying physics and give us inspirations, and to instill existing physical constraints into the model to make the training more efficient~\cite{karniadakis2021physics,liu2021knowledge}.

\begin{figure}[t]
  \centering
   \includegraphics[width=0.9\linewidth]{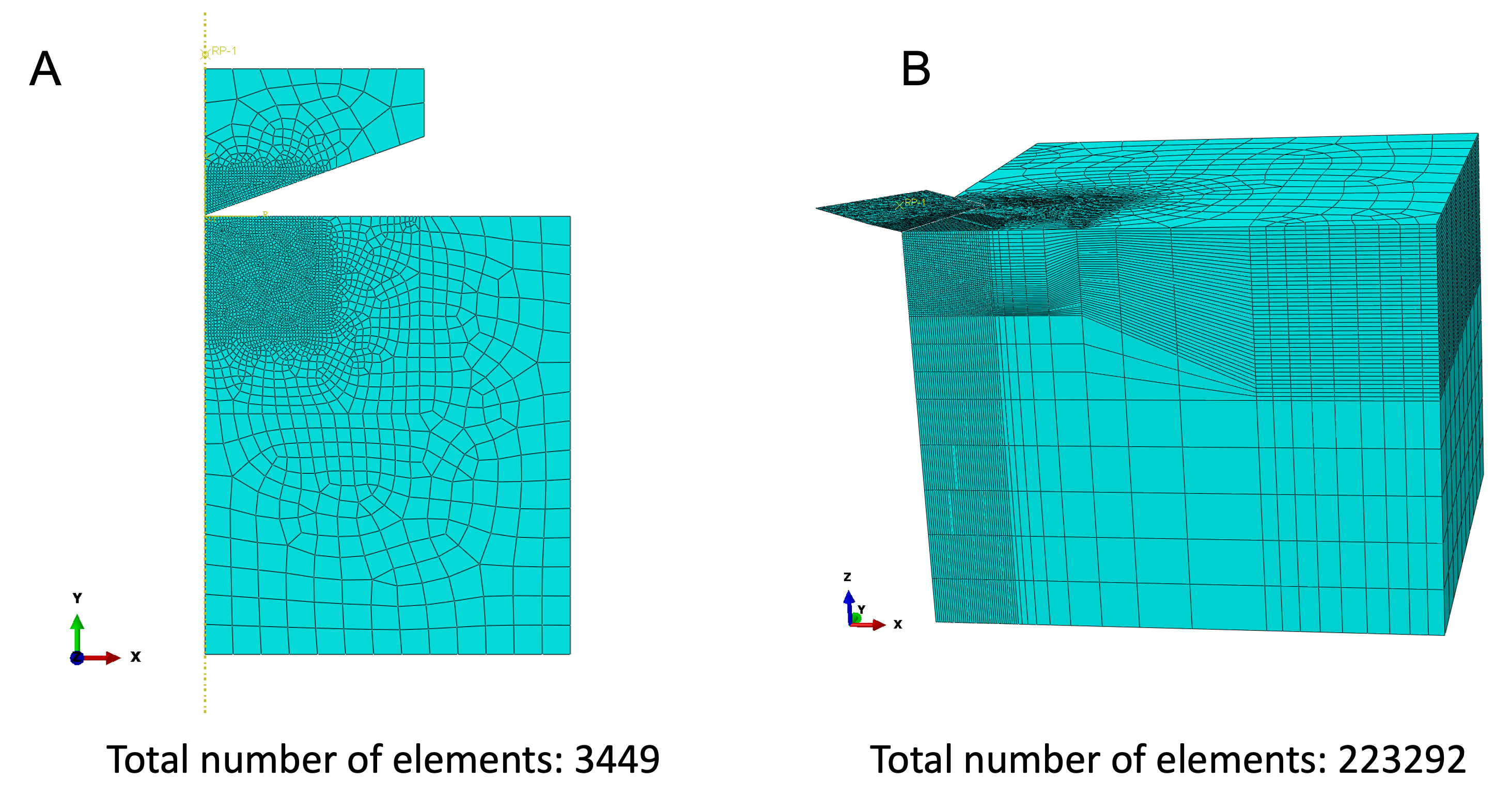}

   \caption{2D and 3D FEM models in our study. The total element number is 3449 in 2D FEM (A) and 223292 in 3D FEM (B).}
   \label{fig:threefigure}
\end{figure}

\begin{figure}[t]
  \centering
   \includegraphics[width=0.9\linewidth]{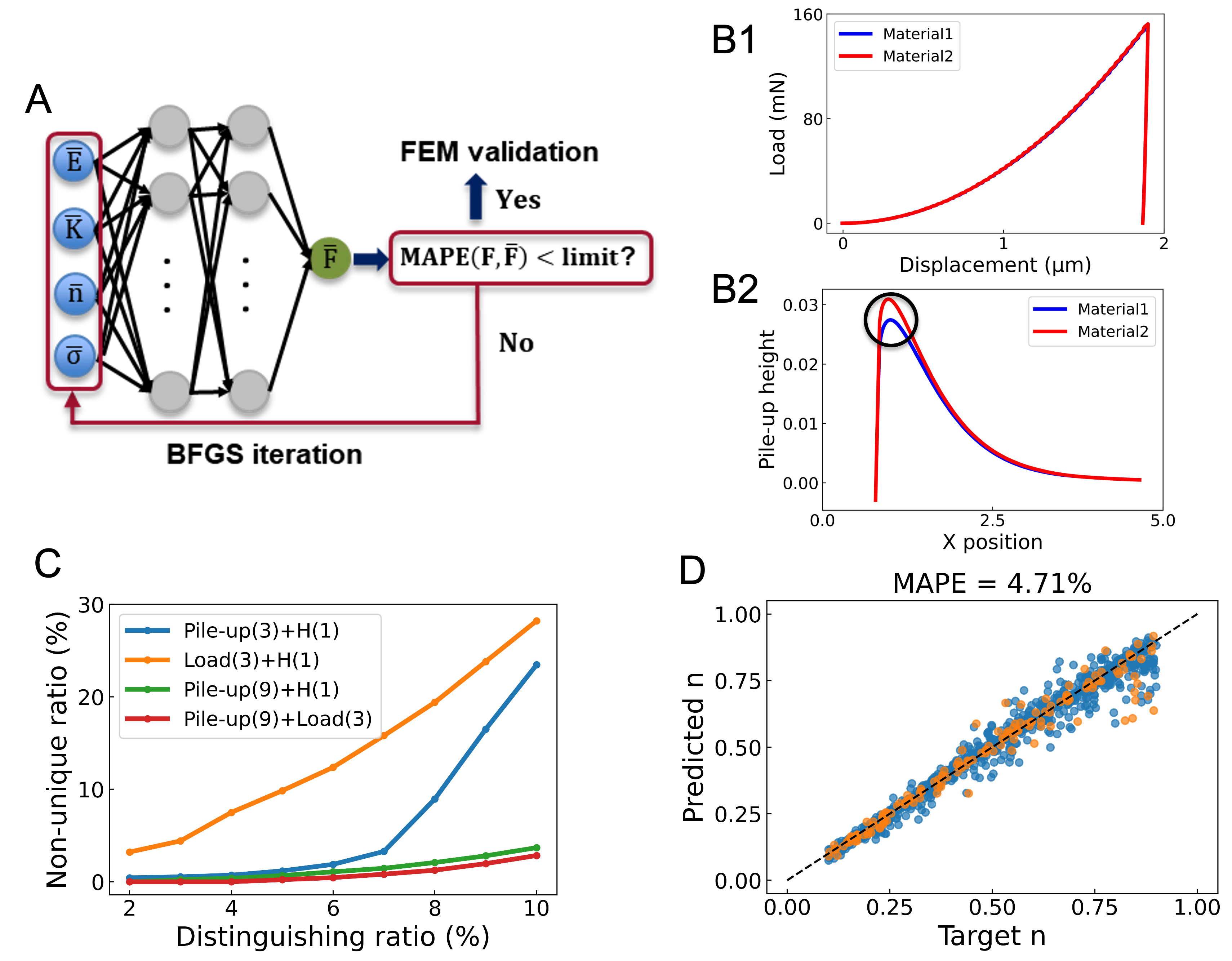}

   \caption{Forward prediction with the unique problem and inverse prediction with the feature selection process. (A) Forward prediction combining BFGS optimization to find the mystical material siblings with the same indentation features. (B1-B2) Two typical material siblings (($E, \sigma_y, n, K$) = (200, 0.28, 0.65, 1.365), (203, 0.254, 0.485, 1.020)) corresponding to almost the same load-displacement curves. (C) Plot of Non-unique ratio \vs Distinguishing ratio. (D) Value correlation of the predicted $n$ and the target $n$. The results are based on 4000 2D FEM data with 3500 training data (blue) and 500 testing data (orange). }
   \label{fig:fourfigure}
\end{figure}

\section{Experiments and results}
\label{sec:expe}
\subsection{Unique problem and feature selection}
One fundamental question is whether the features we choose can guarantee a unique problem. Here we artificially define the feature extraction method based on curve characteristics, as illustrated in Appendix~\ref{sec:appendix3}. Specifically, we can extract three features from force curves based on the the loading curvature, initial unloading slope, and the ratio of residual unloading depth to maximum loading depth. We extract nine features from pile-up curves by finding the maximum height and calculating the volumes and weighted centres in varied parts. We also have tried using encoder-decoder structure to automatically output features, while the testing results show that both types of features perform similarly.

As shown in \cref{fig:fourfigure}A, we use 2D FEM data (\cref{fig:threefigure}A) to build up an accurate forward prediction model, \ie, predicting the force and pile-up features based on input constitutive model parameters. The data number is actively augmented to ensure that the Mean Absolute Percentage Error ($MAPE$), defined as follows,
\begin{equation}
        MAPE = \dfrac{1}{N}\sum_{i=0}^{N} \left|\dfrac{T_i-P_i}{T_i}\right|
\end{equation}
, for predicting each feature is below 2\%. $T_i$ and $P_i$ are the true and prediction values of the $i^{th}$ data point, respectively. Each time 50 data points are added into the parameter range with large errors, and the total data number for forward prediction is 2450. Then we use this trained NN as a surrogate model to explore the informativeness of the indented features. To check whether one material owns siblings with quite similar features, we fix its parameters and iterate the parameters of the candidate material to continually decrease their feature differences. The iteration process uses typical BFGS optimization algorithm. The iteration stops when the $MAPE$ of all features are below the set limit or the iteration number exceeds the upper limit. Finally, we carry out FEM simulations to verify whether the hypothesized material siblings own similar features. \cref{fig:fourfigure}B1 and \cref{fig:fourfigure}B2 display two typical material siblings acquired from the above-mentioned workflow, in which the load-displacement curves are almost the same and the pile-ups reveal some difference at the highest parts.

Theoretically, we can distinguish two materials if the maximum difference among their features exceeds the possible variance, \eg, the experimental errors. Hence, we define the concept of distinguishing ratio as, 
\begin{equation}
        Distinguishing\;ratio = \max_{i=1,2,...,N} \left|\dfrac{F_i-\bar F_i}{F_i}\right|
\end{equation}
, where $F_i$ and $\bar F_i$ are the features from two material siblings, respectively.  $N$ depends on how many features considered. We then take a uniform grid (grid number = 5) on the parameter space of Ludiwk model, and test the uniqueness of 625 materials. Among these 625 materials, $N_A$ of them own material siblings with all the relative feature differences lower than the distinguishing ratio. Then we define the non-unique ratio at this specific distinguishing ratio as $N_A/625$, displaying the possibility to encounter unique problems. \cref{fig:fourfigure}C shows the evolution of the non-unique ratio with the distinguishing ratio employing different features. It reveals that the unique problem is quite severe if only three force features and hardness are input as features. However, the non-uniqueness is largely mitigated if we also include the information of pile-up. Even if we choose only three features from pile-up combined with hardness, the performance is still much better than the pure force case when the distinguishing ratio is lower than 6\%. Meanwhile, we find that the non-unique ratios with nine pile-up features and three force features are close to the case with nine pile-up features and hardness, both observing slight increases when the distinguishing ratio is higher than 8\%. This hints that when including the information of pile-up, maybe we can represent the information of load-displacement relation with only hardness. We then verify this hypothesis by doing the inverse training with nine pile-up features and hardness. The $MAPE$ of each predicted parameter ($E$, $\sigma_y$, $n$, $K$) for the testing set decreases to lower than 5\% when the total data number increases to 4000, as shown in \cref{fig:fourfigure}D. 

\begin{figure*}[t]
  \centering
   \includegraphics[width=0.7\linewidth]{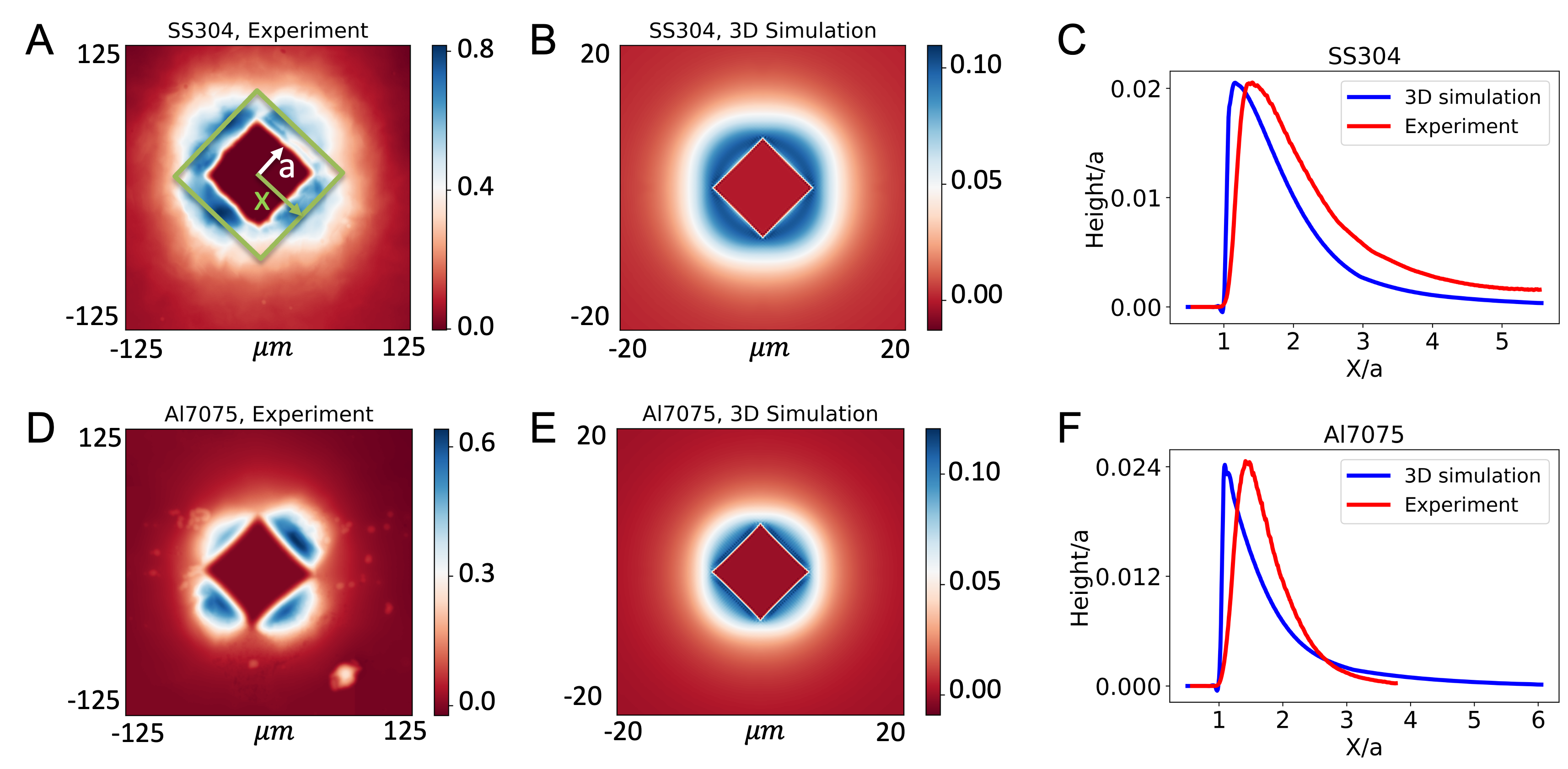}

   \caption{Pile-up profiles acquired from experiments and simulations. (A-C) and (D-F) are pile-up profiles of SS304, and Al7075, respectively. (A) SS304 pile-up profile measured from experiments. The vertical heights of four-fold profiles are divided into slim square strips and averaged \vs the horizontal distance (X, green color) from the origin. Both the heights and distances are normalized by the indentation lateral length (a, white color). All the 3D pile-up profiles (A-B, D-E) are dissolved through this method to form into 2D pile-up curves (C) and (F).}
   \label{fig:fivefigure}
\end{figure*}

\subsection{2D to 3D simulation model transfer}
We then discuss the transfer learning framework of our models. To decrease the burden of expensive data like 3D FEM simulations and experiments, we first build up a baseline model with many pure 2D FEM data, and then ameliorate the gap between 2D and 3D FEM models with some 3D FEM data. Finally, we calibrate the gap between 3D FEM simulations and experiments with several experimental data. The features we use in all the following conditions are based on nine pile-up features and hardness.

The numerical indenter simulated in the 2D axisymmetric FEM modeling is in the conical shape, in which the pile-up morphology can be represented by a one-dimensional (1D) curve, as shown in \cref{fig:fourfigure}B2. However, the indenters used in most hardness testing experiments are fourfold Vickers indenters without the axisymmetric property. \cref{fig:fivefigure}A and \cref{fig:fivefigure}B show two typical 2D images of pile-up morphologies in the experiment and 3D FEM simulation, respectively. Due to surface roughness and grain variations in real metals, the raw pile-up morphologies acquired from experiments are not perfectly smooth in heights (\cref{fig:fivefigure}A). To mitigate the height variations and transform the 2D image into the 1D curve, we divide the 2D image into many slim square strips and calculate the average height in each divided region. Then we acquire a plot of the averaged height \vs the distance to the imprint center and calculate related features, as shown in \cref{fig:fivefigure}C. 

\begin{figure}[t]
  \centering
   \includegraphics[width=0.95\linewidth]{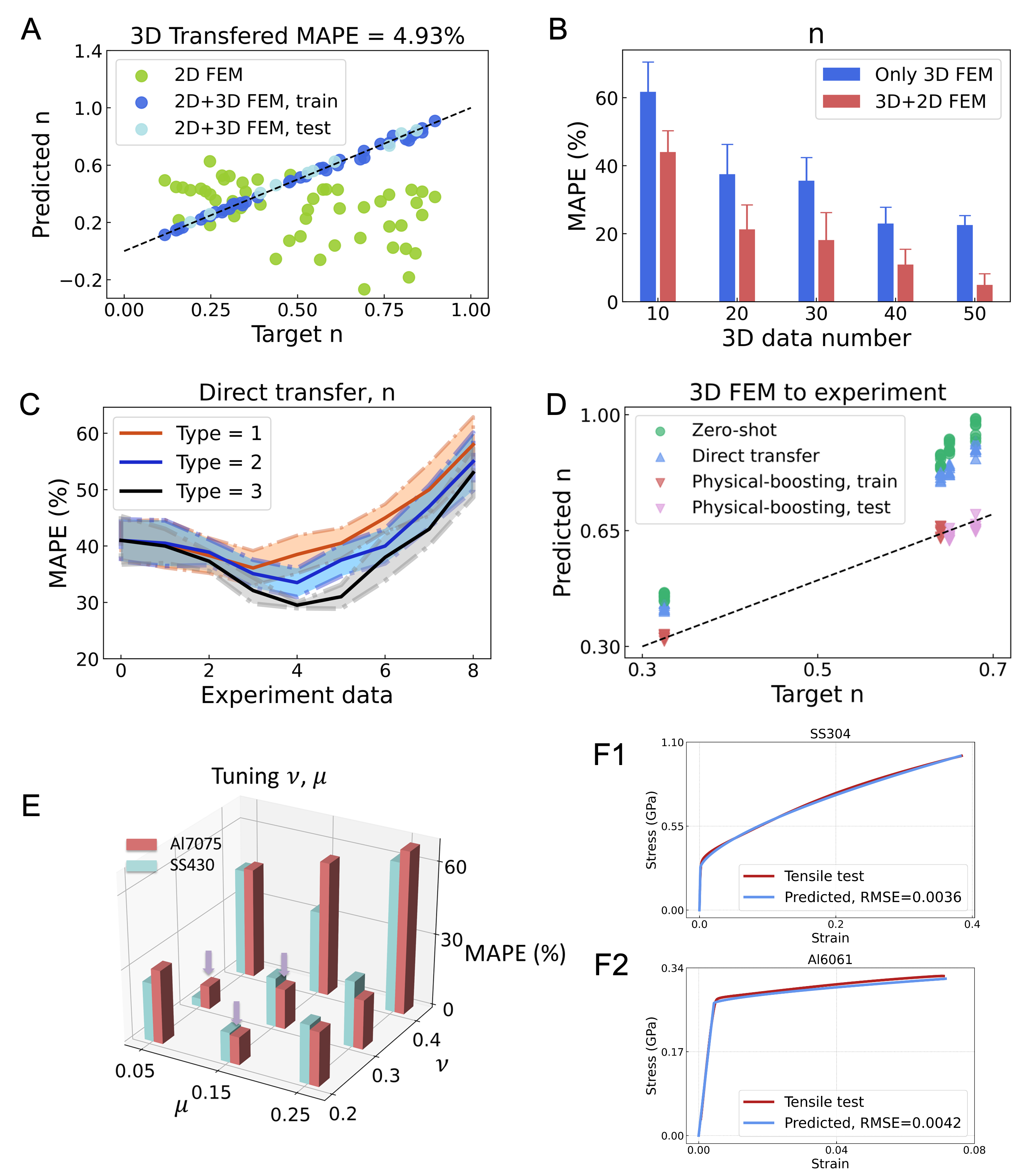}

   \caption{A transfer learning framework for the real experimental prediction. (A) 2D to 3D FEM model transfer with total 50 3D FEM data. The green and blue points are respectively the prediction of $n$ for 3D FEM with pure 2D model and 3D transferred model. (B) MAPE of $n$ \vs 3D FEM data number with (red) or without (blue) 2D baseline model. (C) MAPE of $n$ \vs experiment data number via the method of direct transfer. Here the experiment data number refers to the data points of the same material. The number of types refers to the types of materials used for transfer learning. (D) Comparison of predicting accuracy for three transfer learning methods employed in this study. (E) The choice of physical parameters, \ie, friction coefficient $\mu$ and Poisson’s ratio $\nu$, for physical-boosting. Here the MAPE measures the feature difference between simulations and experiments of Al7075 (red) and SS430 (blue). (F1) and (F2) The ground truth and predicted stress-strain curves of SS304 and Al6061, respectively.}
   \label{fig:sixfigure}
\end{figure}

Since different shaped indenters will result to different pile-up morphologies and corresponding features, using the ML model trained by the 2D FEM data to predict the 3D FEM result will contain large errors. We first use 4000 2D FEM data to build up a 2D ML model and directly employ it for the prediction of 3D FEM results, denoted as $Y1$ in \cref{fig:twofigure}. The green points in Fig. 4A display the prediction of $n$ in this case, and the errors are quite large. However, these predicted Y1 still incorporate the information of 2D FEM data, and they own a rough trend with the true values. To correct the wrong correspondence and transfer the model from 2D to 3D, here we set the predicted $Y1$ as the extra feature and put it together with the original pile-up features and hardness extracted from 3D FEM model to predict the target parameters ($E$, $\sigma_y$, $n$, $K$). \cref{fig:sixfigure}B shows the evolution of MAPE \vs the 3D data number for the case with and without the extra feature ($Y1$). We find the MAPE of the case incorporating 2D prediction will decrease to lower than 5\% when the data number increases to 50, much lower than the case using pure 3D data. \cref{fig:sixfigure}A shows the correspondence of the predicted and targeted values based on this transferred model. The transfer learning framework decreases the requirement of expensive data.
\subsection{Sim-to-real model transfer (Physical-boosting)}
After acquiring a 3D ML model, the next step is to transfer it into the real experimental ML model. However, as shown in \cref{fig:fivefigure}C and \cref{fig:fivefigure}F, the pile-up curves of 3D simulations and experiments are not perfectly consistent, even if the input stress-strain curves for simulations are acquired from the real materials. This inconsistency may be caused by many systematic errors, e.g., the indenter tip is not perfectly sharp, the over-simplified assumption of Poisson’s ratio $\nu$ to be 0.3 and friction coefficient $\mu$ to be 0.15, and the grain effects. The feature value differences between experiments and simulations are larger than 20\% in some features, while the evolution trends with different materials are consistent. The destination target in this section is to use some types of materials for calibrating sim-to-real gap, and then employ this calibrated model to predict the stress-strain relation of other types of materials.

Under the constraints of the limited number of metal types in the real experiments or industries, one intuitive way to mitigate the sim-to-real gap is by incorporating some experimental data points with the 3D FEM data points together, and apply these merged data for the transfer learning process from 2D FEM model to 3D model, the same as the process mentioned in the section 2D to 3D simulation model transfer. This time the 3D model is not based on pure 2D/3D FEM data, since some experimental data are also used. Here in our experiments, each type of material has been measured 8 times repetitively. The number of data points chosen out of these 8 points for the transfer learning should be determined. We then plot the evolution of MAPE with the experiment data number in \cref{fig:sixfigure}C. The number of materials used for the training ranges from 1 to 3 (one-shot to three-shot), respectively. The MAPE are calculated from the other 20 testing materials. The specific choice of material types and experimental data points are randomly repeated for 20 times. With the experiment data number increasing, the MAPE will first decrease and then increase when the experiment data number exceeds 4. This phenomenon is reasonable since an overlarge number of data points of the same material tends to induce overfitting.

According to the above discussion, the transfer learning will be inefficient if the experimental data is rare and the sim-to-real gap is too large. In the above FEM simulations, the Poisson’s ratio $\nu$ and the friction coefficient $\mu$ are always assumed to be fixed values ($\nu$ = 0.3, $\mu$ = 0.15), which are common settings in most indentation models~\cite{chen2007uniqueness,goto2019determining,haj2008nonlinear}. However, our further study reveals that these two physical parameters will greatly impact the acquired features, as illustrated in Appendix~\ref{sec:appendix4}. We vary the $\nu$ to be (0.2, 0.3, 0.4) and the $\mu$ to be (0.05, 0.15, 0.25), and then calculate the evolution of pile-ups in these $3^2$ cases. For each type of features, the maximum difference among these 9 cases will range from 20\% to 40\%. The changing ratios will also vary among different materials. We then calculate the corresponding features of the materials (Al7075, SS430) and compare them to the experimental features. \cref{fig:sixfigure}E displays the total feature differences between simulations and experiments under varied $\nu$ and $\mu$. Among the nine combinations, the sim-to-real gap will be lower than others if the two physical parameters ($\nu$, $\mu$) = [(0.3, 0.15), (0.3, 0.05), (0.2, 0.15)]. Next, we build up three independent DNNs in which the input 3D FEM data are calculated by setting two physical parameters to be the above three values, respectively. Specifically, the parameters ($\nu$, $\mu$) are set as fixed values (0.3, 0.15) for 4000 2D FEM simulations, while changed to [(0.3, 0.15), (0.3, 0.05), (0.2, 0.15)] for each subset of 50 3D FEM data. The total number of 3D FEM data used increases to 150.  

Then for each DNN, using the merged data of 3D FEM and experiments, one candidate value will be predicted, denoted as $Y3$, $Y3'$, $Y3''$ in three independent subsets. These three DNNs form into a committee and finally determine what the optimal prediction value is. The specific weights for each DNN are tuned as, 
\begin{equation}
        Y4=\alpha1*Y1+\alpha2*Y2+(1-\alpha1-\alpha2)*Y3
\end{equation}
Here $Y4$ is the final predicted value in our model and ($\alpha1$, $\alpha2$) are two parameters to be determined by experimental data. To stabilize the training, we use $Sigmoid$ function to constrain the values of ($\alpha1$, $\alpha2$) between 0 and 1. We use the experimental data of SS430 and Al7075 as the training data and find that the training error reaches minimum when $\alpha1=0.19$, $\alpha2=0.33$. Through forming a committee under varied physical parameters and fine-tuning the relative weights, the $\nu$ and $\mu$ represented in simulations can be closer to the experimental condition and other systematic biases such as tip radius effects of nominally sharp indenters can also be actively mitigated. More specifically, the reason why the settings [(0.3, 0.15), (0.3, 0.05), (0.2, 0.15)] are closer to experiments is not necessarily owing to the closer $\nu$ and $\mu$ values as that in experiments, but that these settings possibly offset the systematic biases caused by other factors. This whole process is named as ‘physical-boosting’ since the intuition is based on tuning two physical parameters in the simulations. In summary, the experimental training data has been used three times throughout the ML model, i.e., the choice of physical parameters with three least sim-to-real gaps, combining with 3D simulation data to train DNNs, fine-tuning the relation among three independent DNNs. We use this final model to predict the stress-strain relation of the left 20 other types of materials, serving as the testing set (SS304 (\cref{fig:sixfigure}F1) and Al6061 (\cref{fig:sixfigure}F2) as the examples shown). The predicted stress-strain curves are satisfyingly close to the real curves acquired by tensile testing (the mean relative errors of predicted stresses are 3.4\% across all testing materials). The full accuracy table is in Appendix~\ref{sec:appendix5}.

During forming the committee of three ($\nu$, $\mu$) pairs, the initial grid search for choosing three closest ones is indispensable, since the member with quite large sim-to-real gap will even undermine the model performance.
\subsection{Pointwise stress-strain prediction}
\begin{figure}[t]
  \centering
   \includegraphics[width=0.95\linewidth]{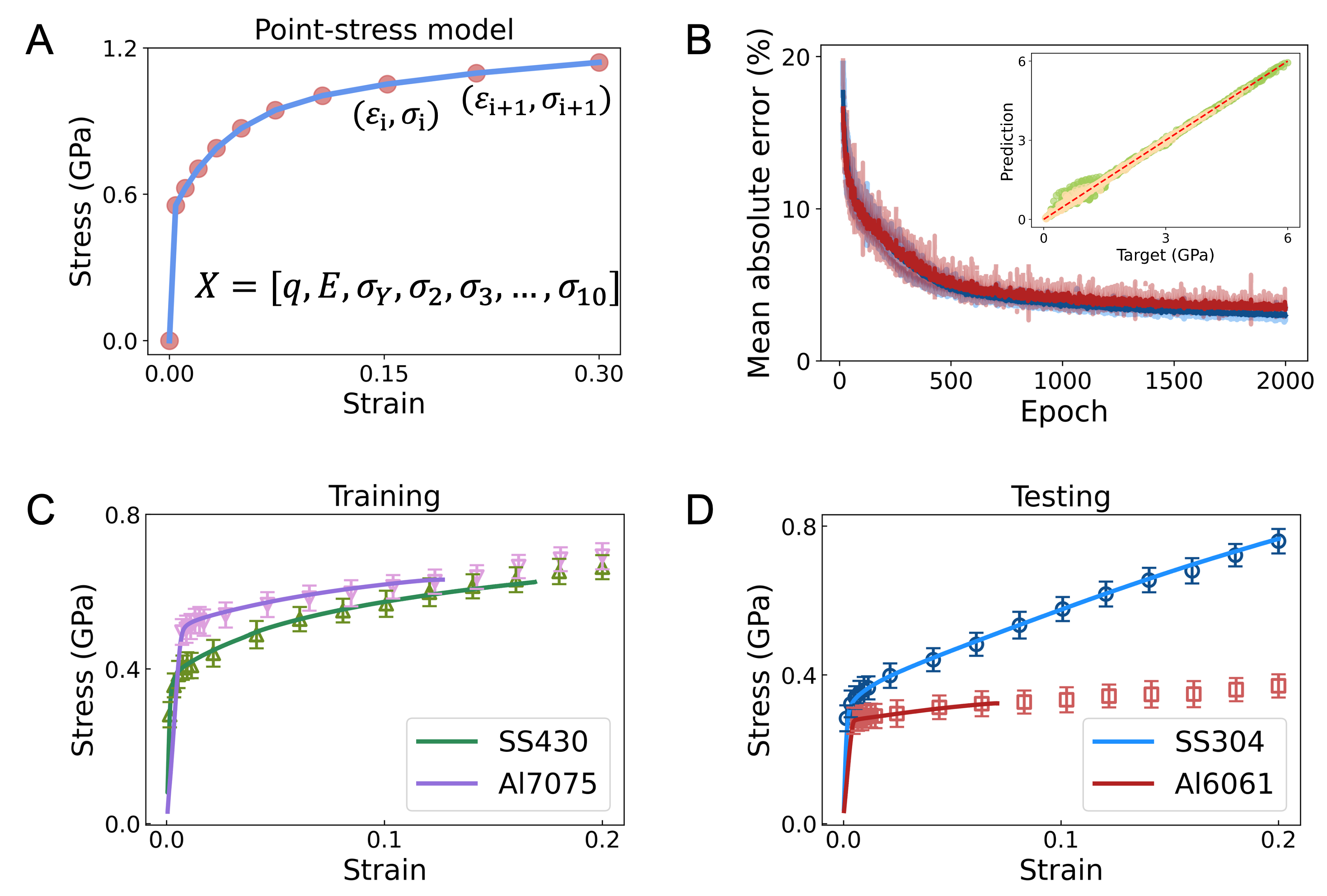}

   \caption{Direct pointwise prediction without the advanced assumption of constitutive models. (A) The created random stress-strain curves conforming to the point-stress model. (B) The red and blue shaded lines represent the MAE over the training epochs for testing and training datasets, respectively. The DNNs are trained 40 times with different initial weights. The inset shows the pointwise comparison between predicted stresses and target stresses. The green and yellow points are training and testing datasets. (C) and (D) The final pointwise prediction from transferred model for training and testing materials, respectively.}
   \label{fig:sevenfigure}
\end{figure}
To make the prediction more tenable and represent a much wider material space, here we add another kind of material model into the training and testing dataset, called ‘point-stress model’. In this model, the linear elastic part is determined by elastic modulus $E$ and yield stress $\sigma_y$, while the nonlinear plastic part is represented by nine strain-stress points, i.e., [$(\varepsilon_i, \sigma_i), i=2,3,…10$], as shown in \cref{fig:sevenfigure}A. The whole tensile curve is estimated between the points by linear interpolation. The yield strain $(\varepsilon_y=\dfrac{\sigma_y}{E})$ and the maximum strain $(\varepsilon_{10}=0.3)$ are then limits of the total interval of the strain values $\varepsilon_i$ to be estimated on the tensile curve. From this interval, the intermediate strain values, representing the positions of the $2^{nd}$ to $9^{th}$ points to estimate, are calculated by adopting a geometric progression. This choice of progression serves to obtain a higher density of points at lower strains, aiming to capture more features near the yield stress. The specific equation is defined as,
\begin{equation}
        \varepsilon_{i+1}-\varepsilon_i=q*(\varepsilon_i-\varepsilon_{i-1}), \;i=2,3,...,9
\end{equation}
Here the common ratio $q$ ranges from 1.1 to 1.5 in different materials. A larger $q$ corresponds to denser aggregation of points near yield stress. After determining strain points $\varepsilon_i$, the corresponding stress values $\sigma_i$ are randomly generated under the constraint of softened-hardening behavior, displayed as,
\begin{equation}
        k_i=\dfrac{\sigma_i-\sigma_{i-1}}{\varepsilon_i-\varepsilon_{i-1}}, \;i=2,3,...,10
\end{equation}
\begin{equation}
        k_{i+1}<k_i,\;k_{i+1}>0, \;i=2,3,...,9
\end{equation}
In summary, the variables to be determined in the point-stress model are [$q,E,\sigma_y,\sigma_2,\sigma_3,…,\sigma_{10}$].

Then the dataset for the pointwise stress-strain prediction of 2D FEM simulations comprises 2000 trials of Ludwik model, 1000 trials of Hollomon model, and 1000 trials of the point-stress model. As for the inverse prediction, we also try to predict the discrete strain-stress points and linearly interpolate them into the whole tensile curve, in which the objective parameters are [$E,\sigma_y,(\varepsilon_1^*,\sigma_1^*),(\varepsilon_2^*,\sigma_2^*),…,(\varepsilon_m^*,\sigma_m^*)$]. Here the total number $m$ of strain-stress points are artificially determined by humans, even can be numerous if the computing resource is enough. Meanwhile, the specific positions of strains can also be flexible, different from the points in the input point-stress model. We set 6 strain positions uniformly gridded on [$\dfrac{\sigma_y}{E}, \dfrac{\sigma_y}{E}+0.01$], and 10 strain positions uniformly gridded on [$\dfrac{\sigma_y}{E}+0.02, 0.2$]. \cref{fig:sevenfigure}B shows the evolution of the average mean absolute errors of all the objective parameters with the iterations (epochs). Both the training error (blue) and the testing error (red) will decrease to less than 5\% when epochs arrive 2000. The inner plot shows the corresponding comparison between the true stress and the predicted stress. The predicting accuracy is magically good, hinting that each part of the stress-strain tensile curve may have its unique influences on the final pile-up morphologies.

The remaining process of transferring the 2D model into the 3D model, and then into the experimental model is the same as that stated in the section: Sim-to-real model transfer (Physical-boosting). For the 3D simulation data, apart from the original 150 trials of Ludwik model and 60 trials of Hollomon model, 60 trials of the point-stress model are added. The final predicted strain-stress points for the two training materials and two testing materials in experiments are displayed in \cref{fig:sevenfigure}C and \cref{fig:sevenfigure}D, respectively. The overall prediction accuracy is satisfying.
\section{Discussion}
Herein, we attempt to bridge the gap between optical residual profiles and material elasto-plastic properties via MFNN. How to use machine learning to explore the inverse problem and how to supplement the physical constraints into the model are discussed. Considering the high cost to get real experiment data, we try to build up a method with only one-shot or few-shot calibrations. The testings results show excellent acuracy. Combining novel imaging and vision techniques with object disturbances is a powerful modality that deserves more attention. This work is a nice demonstration that adding information that can be easily captured by an interferometer significantly improves reconstructions.
\label{sec:disc}

\section*{Impact Statements}
This paper presents the work to connect machine learning with material discoveries using optical image as the information. It will serve as one great example of applying machine learning into scientific research, especially under the constraints of data limitation and fidelity variance.

\bibliography{example_paper}
\bibliographystyle{icml2024}

\newpage
\appendix
\onecolumn
\section{Pre-assumed constitutive models}
\label{sec:appendix1}

\begin{figure}[h]
  \centering
   \includegraphics[width=0.7\linewidth]{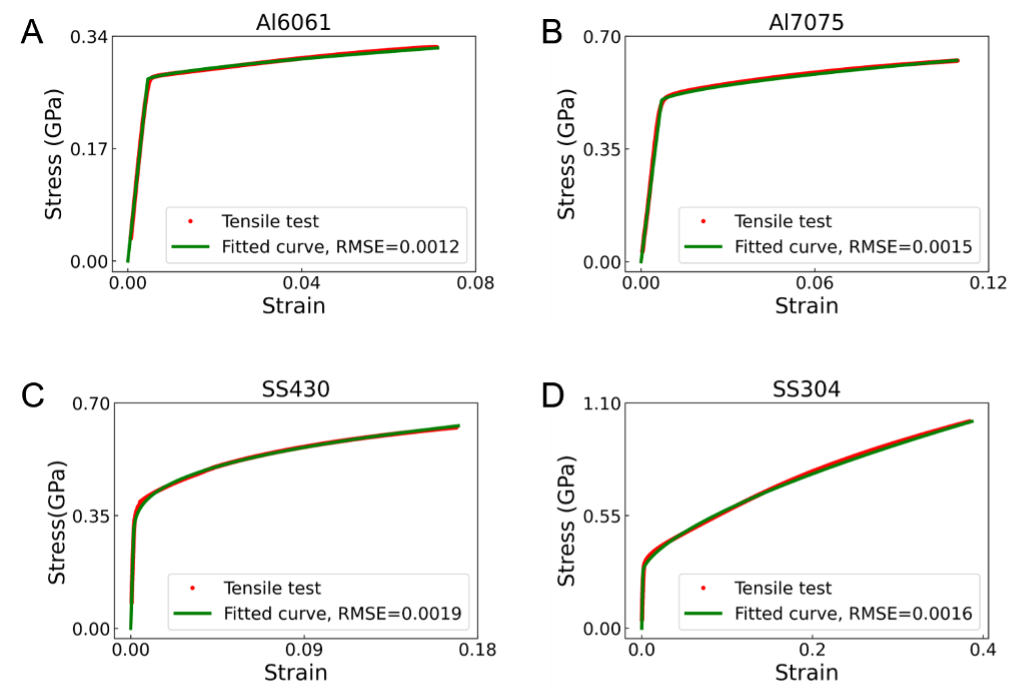}

   \caption{Fitting of experimental tensile curves via four-parameter Ludwik model. The red dots show the experimental results, while the green curves present the fitted curves. Ludwik model can well fit the stress-strain behaviors in all the four metals.}
   \label{fig:appendixfigure1}
\end{figure}

\begin{figure}[H]
  \centering
   \includegraphics[width=0.7\linewidth]{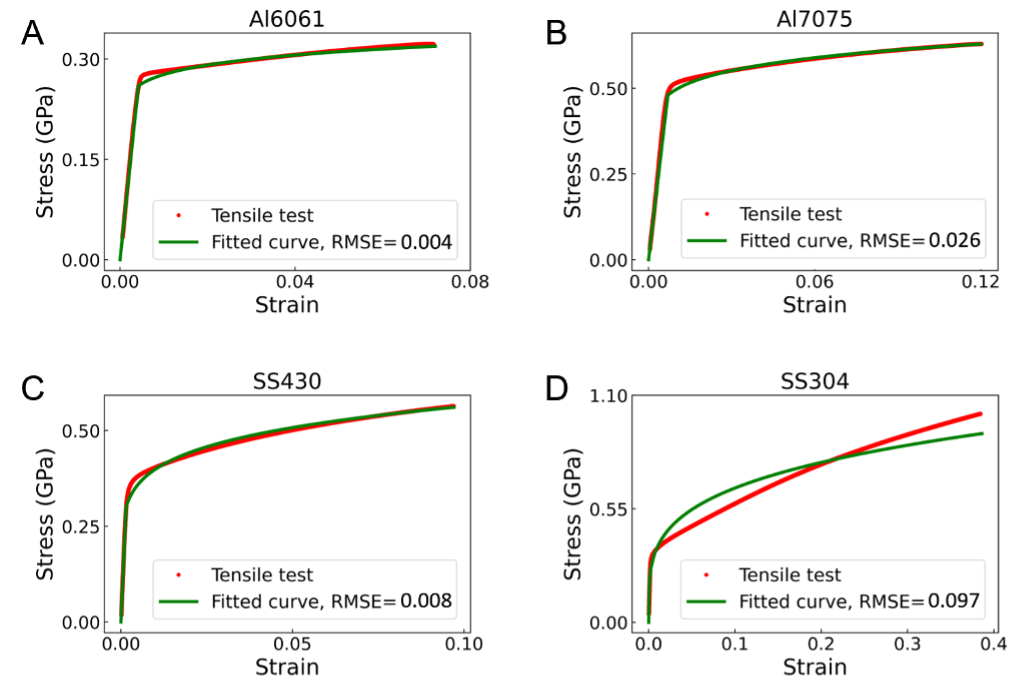}

   \caption{Fitting of experimental tensile curves via three-parameter Hollomon model. The red dots show the experimental results, while the green curves present the fitted curves. Hollomon model cannot well fit the stress-strain behaviors of high-hardening metals, such as the SS304 in this study.}
   \label{fig:appendixfigure2}
\end{figure}

\newpage
\section{Dataset construction}
\label{sec:appendix2}

\begin{table}[h]
\centering
\caption{Parameter ranges of Ludwik/Hollomon model for the dataset of 2D/3D FEM. Here the chosen range of parameters is wide enough to nearly cover all the metallic materials.}
\label{tab:model_parameters}
\begin{tabular}{@{}lcccc@{}}
\toprule
Models    & \( E \) (GPa) & \( \sigma \) (GPa) & \( n \)   & \( K \)   \\ \midrule
Ludwik    & 30--300       & 0.05--1            & 0.1--0.9  & 0.1--2    \\
Hollomon  & 30--300       & 0.05--3            & 0.05--0.5 & N/A       \\ \bottomrule
\end{tabular}
\end{table}

\begin{table}[h]
\centering
\caption{The datasets and sizes used in this study.}
\label{tab:dataset_sizes_2}
\begin{tabular}{@{}lc@{}}
\toprule
Dataset                          & Size  \\ \midrule
2D FEM, Ludwik                   & 4000  \\
2D FEM, Hollomon                 & 1000  \\
2D FEM, Point-stress             & 1000  \\
3D FEM, Ludwik                   & 150   \\
3D FEM, Hollomon                 & 60    \\
3D FEM, Point-stress             & 60    \\
Al6061, experiment               & 8     \\
Al7075, experiment               & 8     \\
Al2011, experiment               & 8     \\
Al3003, experiment               & 8     \\
Al2024, experiment               & 8     \\
Al5052, experiment               & 8     \\
Al6063, experiment               & 8     \\
SS430, experiment                & 8     \\
SS316, experiment                & 8     \\
SS303, experiment                & 8     \\
SS410, experiment                & 8     \\
SS304, experiment                & 8     \\
C26000, experiment                & 8     \\
C22000, experiment                & 8     \\
C71500, experiment                & 8     \\
C10100, experiment                & 8     \\
C11000, experiment                & 8     \\
T1, experiment                & 8     \\
Ti-3Al-2.5V, experiment                & 8     \\
Ti-6Al-7Nb, experiment                & 8     \\
AZ31B, experiment                & 8     \\
AZ91D, experiment                & 8     \\
\bottomrule
\end{tabular}
\end{table}

\newpage
\section{Dataset construction}
\label{sec:appendix5}

\begin{table}[h]
\centering
\caption{Real testing materials and corresponding average relative errors of predicted stresses.}
\label{tab:dataset_sizes}
\begin{tabular}{@{}lc@{}}
\toprule
Testing material                          & Relative error (\%)  \\ \midrule
Al6061               & 3.7     \\
Al2011               & 3.2     \\
Al3003               & 2.1     \\
Al2024               & 4.1     \\
Al5052               & 3.0     \\
Al6063               & 2.7     \\
SS316                & 4.2     \\
SS303                & 2.8     \\
SS410                & 3.4     \\
SS304                & 2.9     \\
C26000                & 3.3     \\
C22000                & 3.6     \\
C71500                & 4.6     \\
C10100                & 4.9     \\
C11000                & 2.0     \\
T1                & 3.0     \\
Ti-3Al-2.5V                & 4.4     \\
Ti-6Al-7Nb                & 2.8     \\
AZ31B                & 3.9     \\
AZ91D               & 5.8     \\
\textbf{Average}              & \textbf{3.4}     \\
\bottomrule
\end{tabular}
\end{table}

\newpage
\section{Feature extraction}
\label{sec:appendix3}
\begin{figure*}[h]
  \centering
   \includegraphics[width=0.9\linewidth]{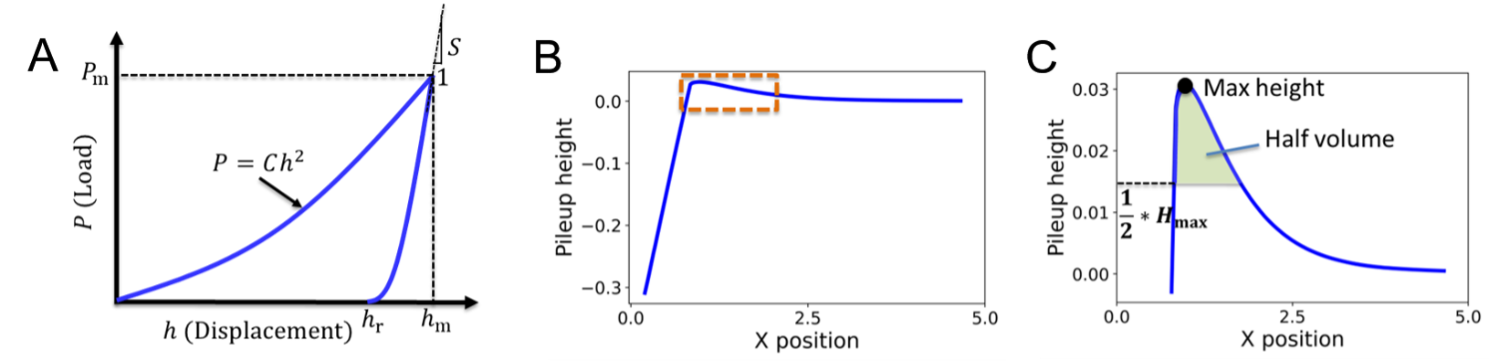}

   \caption{Schematic illustration of chosen features in the load-displacement relation and the pile-up. (A) A typical load-displacement curve from which the loading curvature \(C\), initial unloading slope \(S\), and the ratio of residual unloading depth to maximum loading depth \(h_r/h_m\) are extracted. These three features are mostly used in previous research. (B) A typical 2D pile-up morphology. Here we focus on the part higher than the unindented flattened surface, denoted by the dashed rectangle. (C) Illustration of the features in pile-up. For each pile-up curve, we characterize its maximum height \(H_{\text{max}}\), total pile-up volume \(V_1\), and lateral center coordinate of pile-up \(O_1\). We further find that much information is comprised in the curve part near the position with maximum height. To further extract more information, we only count in the part whose height is higher than \(\frac{1}{2} \times H_{\text{max}}\) and calculate its volume and center coordinate, as shown in the green region. We name these two features half-volume \(V_{1/2}\) and half-center \(O_{1/2}\). In the same way, as for the part with the height higher than \(\frac{3}{4} \times H_{\text{max}}\) and \(\frac{7}{8} \times H_{\text{max}}\), respectively, we extract features fourth-volume \(V_{1/4}\), fourth-center \(O_{1/4}\), eighth-volume \(V_{1/8}\), and eighth-center \(O_{1/8}\). In total, we use 9 features to represent the information of the pile-up.}
   \label{fig:appendixfigure3}
\end{figure*}

\newpage
\section{Impacts of Poisson’s ratio $\nu$ and friction coefficient $\mu$}
\label{sec:appendix4}
\begin{figure*}[h]
  \centering
   \includegraphics[width=0.8\linewidth]{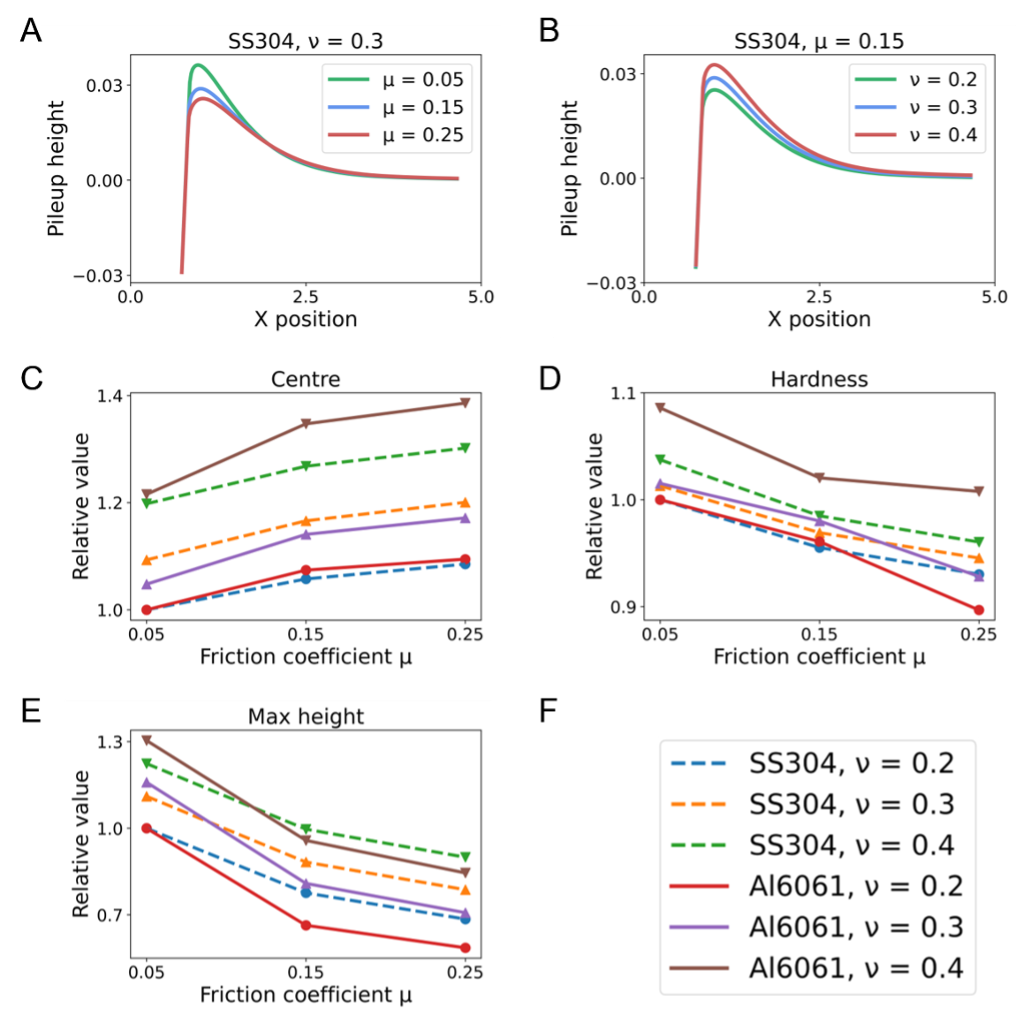}

   \caption{Influence of Poisson's ratio \( \nu \) and friction coefficient \( \mu \) on pile-up profiles. (A) Pile-ups with \( \nu \) fixed to be 0.3, and \( \mu \) varied to be 0.05, 0.15, and 0.25. The pile-up height decreases with the increasing \( \mu \). (B) Pile-ups with \( \mu \) fixed to be 0.15, and \( \nu \) varied to be 0.2, 0.3, and 0.4. The pile-up height increases with the increasing \( \nu \). (C-F) Evolution of the three representative features used in this study (centre \( O_1 \), hardness \( H \), and max height \( H_{\text{max}} \)) with \( \nu \) and \( \mu \). We show the 3D FEM results of SS304 and Al6061 as a comparison.}
   \label{fig:appendixfigure4}
\end{figure*}

\end{document}